\pdfoutput=1
\documentclass[11pt]{article}

\usepackage[]{acl}

\usepackage{times}
\usepackage{latexsym}

\usepackage[T1]{fontenc}
\usepackage[utf8]{inputenc}

\usepackage{microtype}

\usepackage{inconsolata}

\usepackage{amsmath, mathtools}

\usepackage{amssymb}

\usepackage{booktabs}

\newcommand{\spider}{\textsc{spid}er} 

\title{Which Information Matters? Dissecting Human-written Multi-document Summaries with Partial Information Decomposition}
\author{Laura Mascarell \and 
Yan L'Homme \and Majed El Helou \\         ETH Zurich\\
\texttt{lmascarell@inf.ethz.ch}, \texttt{lhommey@ethz.ch}, \texttt{majed.elhelou@inf.ethz.ch}
}
\begin{document}
\maketitle
\begin{abstract}
Understanding the nature of high-quality summaries is crucial to further improve the performance of multi-document summarization. We propose an approach to characterize human-written summaries using partial information decomposition, which decomposes the mutual information provided by all source documents into union, redundancy, synergy, and unique information.~Our empirical analysis on different MDS datasets shows that there is a direct dependency between the number of sources and their contribution to the summary.   
\end{abstract}

\section{Introduction}

% generating English Wikipedia articles can be approached as a multi-document summarization of source documents. 
% abstractive MDS

%However, their model source document should entail the summary proposition. proposition alignment tool to track evidence supporting a certain summary proposition, which only works in the news domain.

Multi-document Summarization (MDS) consists of providing an abridged version of multiple documents.~While some abstractive summarization approaches concatenate all documents into a single input~\cite{johner-etal-2021-error,xiao-etal-2022-primera}, the large size of input text represents a major challenge in MDS. Therefore, most methods implement two-stage approaches that extract salient text spans based on different heuristics~\cite{lebanoff-etal-2018-adapting,j.2018generating,liu-lapata-2019-hierarchical, zhu-etal-2021-twag}.~The text is then fed into a summarization model under the assumption that a high-quality summary is based on such information. While earlier work quantifies the properties of human-written MDS based on n-gram matching~\cite{banko-vanderwende-2004-using}, we argue that there is a lack of in-depth analyses.~Without an understanding of the nature of summaries, improving the quality of MDS remains vague and without clear interpretability.
%of what information from the sources contributes to a high-quality summary. 
%to assess which information humans consider when summarizing multiple documents. 

This work sheds light on what information constitutes a high-quality multi-document summary.~In particular, we propose to categorise the summary information into information provided by at least one source (\textbf{union}) or by a \textbf{unique} source, \textbf{redundant} information from all source documents, and even new information derived from considering them jointly (\textbf{synergy}). In information theory, Partial Information Decomposition (PID) decomposes information in the same way to assess how information about a target is distributed among multiple source variables~\cite{williams2010nonnegative}. 

%This work aims at shedding light on what information constitutes a high-quality multi-document summary, such that more targeted approaches help to improve automatic MDS.

We therefore implement PID in MDS and present \spider; a novel approach to quantify the degree to which the PID components\textemdash such as redundancy or unique information\textemdash contribute to a summary.\footnote{Our implementation also works for single inputs.} We then perform an empirical analysis on human-written summaries from different MDS datasets using our approach. Our results demonstrate that the number of sources has a direct dependence on how they contribute to the summary. We also show that, surprisingly, the order of the source documents matters, and the first three documents are frequently considered as the main source of unique information for any number of sources.

To the best of our knowledge, we present the first fine-grained information analysis in human-written MDS. We open-source \spider\footnote{GitHub:~\href{https://github.com/mediatechnologycenter/SPIDer}{mediatechnologycenter/SPIDer}} and hope that it helps to enhance the performance of future MDS methods. We suggest that our PID approach could also be used to automatically build MDS datasets that align with human quality in future work.

% future works include applying PID for MDS. The framework could potentially be used as a quality measure of MDS quality or to improve MDS.

\section{Information Theory Background}
Mutual Information (MI)~\cite{shannon1948mathematical} has been widely used in NLP tasks to quantify the information that a source provides about an output~\cite{li-etal-2016-diversity,li2016mutual,takayama-arase-2019-relevant,padmakumar-he-2021-unsupervised,mascarell-etal-2023-entropy}.
%also approximated by a language model. 
%Several NLP approaches whose output $Y$ is dependent on a source $X$ use mutual information~\cite{shannon1948mathematical}~$I(X;Y)$ to measure the amount of information that the source provides about the generated output~\cite{li-etal-2016-diversity,takayama-arase-2019-relevant,padmakumar-he-2021-unsupervised,mascarell-etal-2023-entropy}. 
However, mutual information is insufficient in the context of MDS, as it can only be applied to pairs of random variables. 

\begin{figure}
    \centering
    \includegraphics[scale=.034]{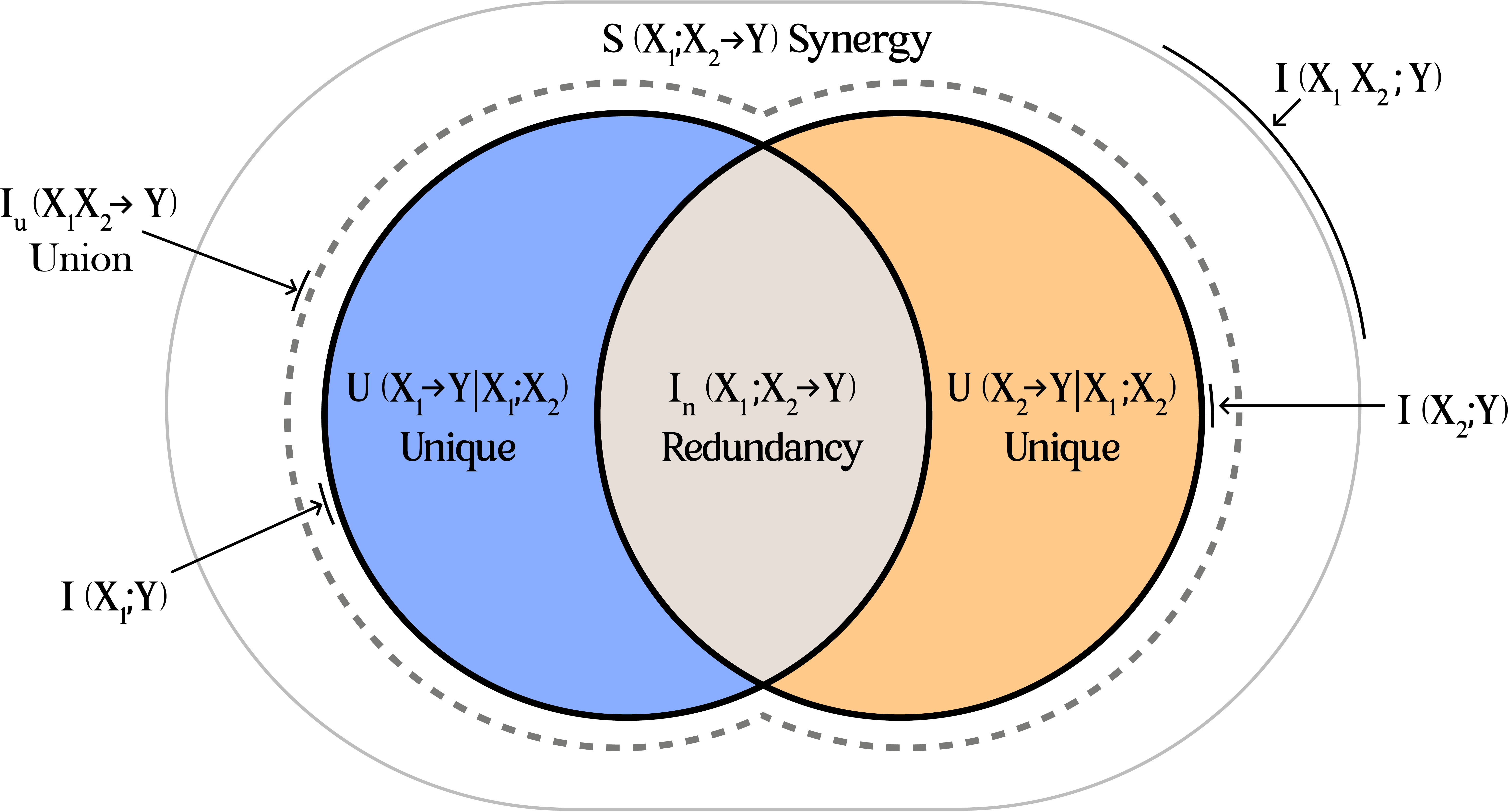}
    \caption{Relationship between the PID components \textit{union}, \textit{redundancy}, \textit{synergy}, and \textit{unique} information that two sources $X_1$ and $X_2$ provide about a target $Y$. $I(X_1,X_2; Y)$ represents the information that both sources provide jointly about $Y$, whereas $I(X_1;Y)$ and $I(X_2;Y)$ represent the information that each source provides individually.}
    \label{fig:pid_plot}
\end{figure}

Partial Information Decomposition (PID) tackles the multivariate problem and decomposes the information that a set of sources conveys about a target into union, redundancy, unique, and synergistic information 
%as follows: information provided by at least one source (\textit{union}) or by a \textit{unique} source, information present in multiple sources (\textit{redundancy}), and \textit{synergistic} information that derives from considering the sources jointly
~\cite{williams2010nonnegative}. To date, PID has only been applied in NLP to measure morphological fusion in~\citet{socolof-etal-2022-measuring}. Figure~\ref{fig:pid_plot} illustrates the relationships between the different PID components for two sources, $X_1$ and $X_2$, and a target $Y$. See a textual example in  Table~\ref{tab:pid-examples}.

In this work, we consider the PID approach proposed in \citet{kolchinsky_pid}, which is based on the definitions of union and intersection from set theory. In contrast to previous work, \citet{kolchinsky_pid}'s approach can be applied to any number of sources, a crucial requirement in our MDS setting.
%Additionally, it allows us to obtain the resulting text of each component via the joint distributions $P_{QY}$. 

Formally, given a set of source random variables $\mathcal{X} = \{X_1,...,X_n\}$ and a target $Y$, \emph{redundancy} $I_{\cap}(\mathcal{X} \to Y)$ is the intersection of information among the sources about $Y$. This redundancy is the maximum information $Q$ we can obtain about $Y$ that is less informative than any of the sources: 
\begin{equation}
%\small
\label{eq:redundancy}
I_{\cap}(\mathcal{X} \to Y) \coloneqq \sup_{Q} I(Y;Q) ~|~ \forall i \text{, } Q \sqsubset X_i%\text{, }
\end{equation}
where $\sqsubset$ is an ordering relation that determines when $X_i$ is more informative than
%\majed{-or equally informative as-} 
$Q$.~Conversely, \emph{union} $I_{\cup}(\mathcal{X} \to Y)$ is the minimum information $Q$ we can obtain about the target $Y$ that is more informative than any of the sources:
\begin{equation}
%\small
\label{eq:union}
I_{\cup}(\mathcal{X} \to Y) \coloneqq \inf_{Q} I(Q;Y) ~|~ \forall i \text{, } X_i \sqsubset Q %\text{.}
\end{equation}
Finally, \emph{unique} and \emph{synergistic} information are derived from Eq.\ (\ref{eq:redundancy}) and Eq.\ (\ref{eq:union}), respectively.
%, as $S(\mathcal{X} \to Y) = I(Y;\mathcal{X}) - I_{\cup}(\mathcal{X} \to Y)$ and $U(Q_i \to Y | \mathcal{X}) = I(Y;Q_i) - I_{\cap}(\mathcal{X}\to Y)$.

\begin{table}
    \small
    \centering
    \resizebox{\columnwidth}{!}{
    \begin{tabular}{rl}
    %\hline 
    \toprule
    %\textbf{Source} \\ 
    %\midrule
    Source $X_1$ & Kimchi is fermented cabbage.\\
    %\midrule
    %\bottomrule
    %\textbf{Summary sentences}\\
    %\midrule
    Source $X_2$ & Fermented foods are rich in probiotics.\\%\midrule
    Target $Y$ & Fermented foods, such as Kimchi, are rich in probiotics.\\\midrule
    Unique $X_1$ & The nature and preparation of kimchi.\\
    Unique $X_2$ & A general characteristic of fermented foods.\\
    Redundancy & Information about \textit{fermented}.\\
    Synergy & Inferring that Kimchi is a fermented food.\\
    \bottomrule
    \end{tabular}}
    \caption{\label{tab:pid-examples} Example with two sources, $X_1$ and $X_2$, and the PID information that they provide about $Y$.}
\end{table}

\section{Decomposing Information in MDS}
We adopt the PID approach in \citet{kolchinsky_pid} to MDS, considering sentences as units of information.~Formally, let $\mathcal{X} = \{D_1,...,D_n\}$ be a set of $n$ source documents, where each document is a collection of sentences $D_i = \{d_i^1,...,d_i^{|D_i|}\}$, and a multi-document summary of $m$ sentences $S = \{s^1,...,s^m\}$. Using Eq.\ (\ref{eq:redundancy}) and (\ref{eq:union}), we define \textit{redundancy} and \textit{union} in MDS as follows:
%Following \citet{kolchinsky_pid}, we define \textit{redundancy} $I_{\cap}^{\textsc{mds}}$ as the maximum information about $S$ in any collection of document sentences $Q$ that is less informative than any of the source documents: 
\begin{equation}
%\small
\label{eq:redundancy_mds}
I_{\cap}^{\textsc{mds}}(\mathcal{X} \to S) \coloneqq \sup_{D \in \mathcal{D}} I(S;D) | \forall i \text{, } D \sqsubset D_i%\text{,}
\end{equation}
%we consider sentences as units of information
% sentences as unit of information
%the optimization set to all possible combinations of sentences contained in the documents $\mathcal{Q} \coloneqq \mathcal{P}(\cup_{i=1}^n Q_i)$.\footnote{This choice implies a finite optimization problem with a set size given by $|\mathcal{Q}| = 2^{\sum_i |Q_i|}$.}
%Similarly, \textit{union} $I_{\cup}^{\textsc{mds}}$ is the minimum information in any collection of sentences that is more informative than any source document:
\begin{equation}
%\small
\label{eq:union_mds}
I_{\cup}^{\textsc{mds}}(\mathcal{X}\to S) \coloneqq \inf_{D \in \mathcal{D}} I(S;D) | \forall i \text{, } D_i \sqsubset D%\text{,}
\end{equation}
where $D = \{d^1,...,d^{|D|}\}$ is a collection of document sentences from all possible sentences $\mathcal{D}$,\footnote{Since optimizing over all collections of sentences is computationally intractable, we implement two optimization strategies. First, we assume that a summary should only contain information from the sources. Therefore, we restrict the set of all possible sentences to the sentences from the sources $\mathcal{D} \coloneqq \mathcal{P}(\cup_{i=1}^n D_i)$, which results in a finite optimization problem with set size $|\mathcal{D}| = 2^{\sum_i |D_i|}$. Second, since the growth of $|\mathcal{D}|$ is still exponential, we implement beam search to find an approximation of the optimal collection of sentences for union and redundancy.} $I(S;D)$ represents the MI between the summary sentences $S$ and $D$, and $\sqsubset$ is an ordering relation. We formally define the ordering $\sqsubset$ and MI in our MDS setting later in this section.

% Optimization decisions: beam search, to find approximately the best collection of sentences for the redundancy and the union information.

We then define \textit{unique} information and \textit{synergy} using Eq.\ (\ref{eq:redundancy_mds}) and (\ref{eq:union_mds}) as in \citet{kolchinsky_pid}
\begin{equation}
%\small
    U^{\textsc{mds}}(D_i \to S | \mathcal{X}) = I(S;D_i) - I_{\cap}^{\textsc{mds}}\text{,}
\end{equation}
\begin{equation}
\label{eq:synergy}
%\small
S^{\textsc{mds}}(\mathcal{X} \to S) = I(S;\mathcal{X}) - I_{\cup}^{\textsc{mds}}\text{,}
\end{equation}
where $I(S;D_i)$ and $I(S;\mathcal{X})$ are the MI between the sentences of a summary and a specific document $D_i$ or all source documents $\mathcal{X}$, respectively.
 
%In the rest of this section, we provide the definition of mutual information $I(S;D)$ and the ordering relation $\sqsubset$ in our MDS setting. We use a language model to estimate all probabilities.\footnote{GPT-2: openai-community/gpt2-large}
%as $S(\mathcal{X} \to Y) = I(Y;\mathcal{X}) - I_{\cup}(\mathcal{X} \to Y)$ and $U(Q_i \to Y | \mathcal{X}) = I(Y;Q_i) - I_{\cap}(\mathcal{X}\to Y)$

% https://huggingface.co/openai-community/gpt2-large
\paragraph{Pairwise Mutual Information}  $I(S;D)$ quantifies the mutual information between summary sentences $S$ and a collection of source sentences $D$. To compute $I(S;D)$, we measure pairwise mutual information on all summary-source sentence pairs $pmi(s;d) = log\frac{p(s;d)}{p(s)p(d)}$ as in \citet{padmakumar-he-2021-unsupervised}, using a language model to estimate probabilities.\footnote{GPT-2 checkpoint: \href{https://huggingface.co/openai-community/gpt2-large}{openai-community/gpt2-large}} To compute $p(s;d)$ the sentences $s$ and $d$ are concatenated and their joint probability is estimated with the language model. We define $I(S;D)$ as the expected value of all $pmi(s;d)$, which is the sum of all values weighted by their likelihood,
\begin{equation}
%\small
    I(S;D) = \mathop{\mathbb{E}}\limits_{s,d \sim S,D}[pmi(s;d)]
   % I(S;D) = \sum_{s \in S} \sum_{d \in D} p_{SD}(s;d) \log \frac{p_{SD}(s;d)}{p_{S}(s)p_{D}(d)}\text{,}
\end{equation}

% stands for the concatenation of q and s.
\paragraph{Ordering Relation}  $D \sqsubset D'$ specifies that a collection of source sentences $D'$ is more informative than a different collection $D$. That is, the information that $D$ provides to a summary $S$ is contained within the information that $D'$ provides to $S$. Hence, we define our ordering relation as:
\begin{equation}
\label{eq:ordering}
%\small
    \begin{split}
        D \sqsubset D' \iff I(S;D) \leq I(S;D') \\ 
        \land \forall s \in S: I(S;D)_{s} \leq I(S;D')_{s}%\text{,}
    \end{split}
\end{equation}
where $I(S,D)_{s} = \mathbb{E}_{d \sim D}[pmi(s;d)]$. Note that the second component of Eq.\ (\ref{eq:ordering}) guarantees that all the information provided by the individual sentences in $D$ is also contained in $D'$.
% beam search & Optimization

\section{PID of Human-written Summaries}
We analyze the information components of human-written summaries from multiple MDS datasets using our framework. Specifically, we consider a random sample of 100 instances per dataset and number of sources, if available (see Table~\ref{tab:datasets}).\footnote{Samples with less than 100 instances: MultiNews with 9 (89 instances) and 10 sources (33); WikiNewsSum with 8 (89 instances), 9 (77), and 10 sources (51).}
%We consider different number of sources if available, and randomly sample 100 instances of each for comparison \textbf{TODO}. 
%Table~\ref{tab:datasets} shows an overview of the analysed data.
%\footnote{100 instances and ones without enough the ones we didn't have enough instances.}

\subsection{Datasets}
% table with avg sentences total summary/sources / number of sources

%mtc/DUC2004_fixedMDS_dataset__sources_10_PID.pkl
%Number of samples: 200
%======================
%Documents sentences
%Mean: 255.26 Max: 574 Min: 147
%Documents
%Mean: 10.00 Max: 10 Min: 10
%Summary sentences
%Mean: 6.90 Max: 16 Min: 3

%mtc/WCEP-filtered_fixedMDS_dataset__sources_10_sample_PID.pkl
%Number of samples: 100
%======================
%Documents sentences
%Mean: 200.20 Max: 876 Min: 60
%Documents
%Mean: 10.00 Max: 10 Min: 10
%Summary sentences
%Mean: 3.24 Max: 6 Min: 2

%% multinews
% summary avg: 13.825925925925926
% doc avg: 32.651028988714316

%% WikiNewsSum
% summary avg: 17.503059975520195
% doc avg: 21.702925243770313

\begin{table}
\small
\centering
\begin{tabular}{lrr|r}
%\hline 
\toprule
%\midrule
\textbf{Dataset} & \textbf{Source} & \textbf{Summary} & \textbf{\#Sources}\\
\midrule
MultiNews & 32.7 & 13.8 & from 2 to 10 \\
WikiNewsSum & 21.7 & 17.5 & from 2 to 10 \\
DUC2004 & 25.5 & 6.9 & 10\\
WCEP & 20.0 & 3.2 & 10\\
%\midrule
\bottomrule
\end{tabular}
\caption{\label{tab:datasets} Overview of the analyzed data. Average sentence count per summary and per source document, and the number of sources considered in each dataset.}
\end{table}

\begin{figure*}
    \centering
    \includegraphics[scale=0.397]{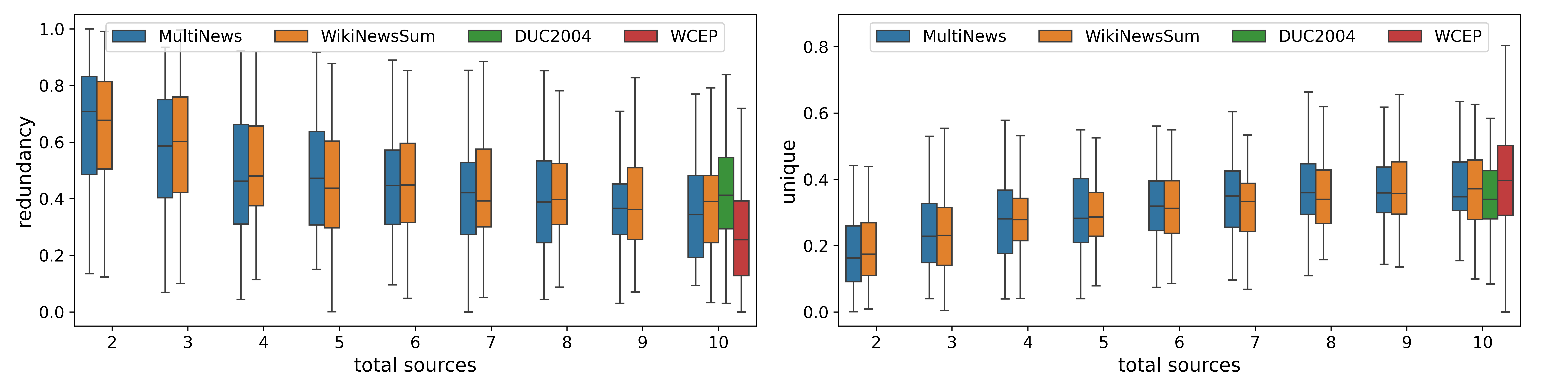}
    \caption{Redundancy (left) and unique (right) information scores across datasets and number of sources. The more sources, the less redundancy and the more unique information contributes to the summary. WCEP scores differ the most from the other datasets. Note that WCEP is extended with additional sources not considered in the summaries.}
    \label{fig:redundancy_plot}
\end{figure*}

\label{sec:final_dataset}

\paragraph{MultiNews} MDS dataset of news articles from the website newser.com, where summaries are written by professional editors. The sources are the cited articles in each summary~\cite{fabbri-etal-2019-multi}. 

\paragraph{WikiNewsSum} The summaries are full articles from the collaborative news platform Wikinews.org and the sources are the cited references. Sources are at least 1.5 times larger than its summary on a character basis \cite{calizzano-etal-2022-generating}.

\paragraph{DUC2004} This dataset consists of 50 sets of 10 articles from the Associated Press and New York Times newswires, each paired with four hand-written summaries. Humans were required to summarize each source independently before writing a multi-document summary of up to 665 characters.\footnote{\url{https://duc.nist.gov/duc2004/}} 
%\footnote{\href{https://duc.nist.gov/duc2004/t1.2.summarization.instructions}{DUC2004 summarization instructions.}}
%This dataset from the Document Understanding Conference 

\paragraph{WCEP} The dataset comprises summaries of news events from the Wikipedia Current Events Portal and the corresponding cited sources (only 1.2 on average). The set of sources are complemented with related articles from the Common Crawl archive~\cite{gholipour-ghalandari-etal-2020-large}, which results in a rather synthetic MDS dataset. Summaries are short, up to 30-40 words; hence, we only consider a subset consisting of multi-sentence summaries of 10 sources each. 

\subsection{Results}
\label{sec:analysis}

\begin{figure}
    \centering
    \includegraphics[scale=.36]{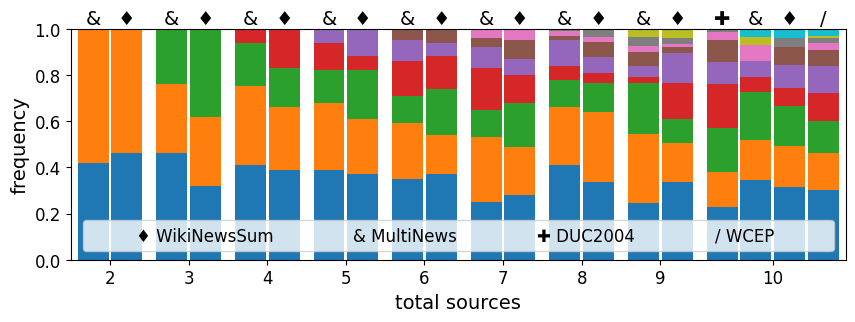}
    \caption{Frequency of each source contributing the most to the summary with their unique information across datasets and total number of sources. The first three sources (blue, orange, and green) contribute the most for any number of sources in all datasets.}
    \label{fig:top_unique_plot}
\end{figure}

%MultiNews
%Union -- Mean: 0.9987393921554724, Median: 1.0, Std_dev: 0.022995864329611093
%Synergy -- Mean: 0.001260607844527652, Median: 0.0, Std_dev: 0.022995864329611093
%Redundancy -- Mean: 0.4752158936756812, Median: 0.4581475666186544, Std_dev: 0.21772103747003427
%Unique total -- Mean: 0.302385734026557, Median: 0.3033796380209239, Std_dev: 0.13555154081688803

%wikiNewsSum
%Union -- Mean: 0.9991143837914159, Median: 1.0, Std_dev: 0.009418696273201065
%Synergy -- Mean: 0.0008856162085840846, Median: 0.0, Std_dev: 0.009418696273201063
%Redundancy -- Mean: 0.4761621105832692, Median: 0.4589424668684101, Std_dev: 0.20829232882795293
%Unique total -- Mean: 0.2966871528919253, Median: 0.2927599287175105, Std_dev: 0.12468190550223579

%% DUC2004
%Union -- Mean: 1.0, Median: 1.0, Std_dev: 0.0
%Synergy -- Mean: 0.0, Median: 0.0, Std_dev: 0.0
%Redundancy -- Mean: 0.4257721714831941, Median: 0.4121327811237886, Std_dev: 0.16794091451610157
%Unique total -- Mean: 0.3456437184452451, Median: 0.33963511828981074, Std_dev: 0.11085708904013516

%mtc/WCEP-filtered_fixedMDS_dataset__sources_10_sample_PID.pkl
%Union -- Mean: 1.0, Median: 1.0, Std_dev: 0.0
%Synergy -- Mean: 0.0, Median: 0.0, Std_dev: 0.0
%Redundancy -- Mean: 0.2854833051148776, Median: 0.2550265718981603, Std_dev: 0.20178780230112633
%Unique total -- Mean: 0.4086240354744505, Median: 0.39665831790045125, Std_dev: 0.1593851079273013

\begin{table}
\small
\centering
\resizebox{\columnwidth}{!}{
\begin{tabular}{lrrrr}
%\hline 
\toprule
%\midrule
\textbf{Dataset} & \textbf{Union} & \textbf{Synergy} & \textbf{Red.} & \textbf{Unique}\\
\midrule
%MultiNews & 1.0 / 0.0 & 0.0 / 0.0 & 0.46 / 0.2 & 0.30 / 0.1\\
%WikiNewsSum & 1.0 / 0.0 & 0.0 / 0.0 & 0.46 / 0.2 &  0.29 / 0.1\\
%DUC2004 & 1.0 / 0.0 & 0.0 / 0.0 & 0.41 / 0.2 & 0.34 / 0.1\\
%WCEP & 1.0 / 0.0 & 0.0 / 0.0 & 0.26 / 0.2 & 0.40 / 0.2\\
MultiNews & 1.0 ($\pm$0.0) & 0.0 ($\pm$0.0) & 0.48 ($\pm$0.2) & 0.30 ($\pm$0.1)\\
WikiSum & 1.0 ($\pm$0.0) & 0.0 ($\pm$0.0) & 0.48 ($\pm$0.2) &  0.30 ($\pm$0.1)\\
DUC2004 & 1.0 ($\pm$0.0) & 0.0 ($\pm$0.0) & 0.43 ($\pm$0.2) & 0.35 ($\pm$0.1)\\
WCEP & 1.0 ($\pm$0.0) & 0.0 ($\pm$0.0) & 0.29 ($\pm$0.2) & 0.41 ($\pm$0.2)\\
%\midrule
\bottomrule
\end{tabular}
}
\caption{\label{tab:analysis} Overall \spider~scores across datasets (mean and standard deviation).}% See all statistics in Table~\ref{tab:apx_analysis}.}
\end{table}

%multiNews
%0.039007584803329304 & 0.0408488260807132 & 0.04972119256947203 & 0.04762399271162211 & 0.048255326400673766 & 0.049677178014347075 & 0.052860114793528944 & 0.050991131031281625 & 0.05588457868832547
%wikiNews
%0.037784902319753136 & 0.043141047825833015 & 0.046052637613996826 & 0.05111883445885161 & 0.04725636228969279 & 0.04791955075904774 & 0.05083133669289949 & 0.05074954223388377 & 0.05101171538548727
%duc 0.04504079653933247
%wcep 0.06464972222239639

\begin{table}
\small
\centering
\resizebox{\columnwidth}{!}{
\begin{tabular}{lrrrrrrrrr}
%\hline 
\toprule
%\midrule
\textbf{Dataset} & \textbf{2} & \textbf{3} & \textbf{4} & \textbf{5} & \textbf{6} & \textbf{7} & \textbf{8} & \textbf{9} & \textbf{10}\\
\midrule
MNews & 3.9 & 4.1 & 4.9 & 4.8 & 4.8 & 5.0 & 5.3 & 5.1 & 5.6\\
WSum & 3.8 & 4.3 & 4.6 & 5.1 & 4.7 & 4.8 & 5.1 & 5.1 & 5.1\\
DUC & - & - & - & - & - & - & - & - & 4.5\\
WCEP & - & - & - & - & - & - & - & - & 6.5\\
%\midrule
\bottomrule
\end{tabular}
}
\caption{\label{tab:unique_variance} Variance of unique information (\%) across datasets and different numbers of sources.}
%The variance increases with the number of sources, indicating higher variability in their individual contribution to the summary.}
\end{table}

We compute \spider~scores on all samples and compare the results among datasets and number of sources. Due to length differences in both summary and sources across datasets, we normalise all values by the total mutual information between the summary and the source documents $I(S;D_1,...,D_n)$. 

%Table~\ref{tab:analysis} shows the overall \spider~scores. 
We observe in Table~\ref{tab:analysis} that \textit{synergy} is negligible in all datasets, and therefore, \textit{union} represents the total mutual information (see Eq.\ (\ref{eq:synergy})). Figure~\ref{fig:redundancy_plot} compares \textit{redundancy} and \textit{unique} information across different numbers of sources.
The results show that redundancy decreases with the number of sources, while the unique information increases. That is, the more sources, the more they contribute individually to the summary. Additionally, DUC2004 scores are comparable to MultiNews and WikiNewsSum with 10 sources, whereas WCEP shows significantly lower redundancy and higher unique information.~WCEP is extended with Common Crawl articles, so it is unclear how these additional articles should contribute to the summary in a MDS task. The difference in \spider~scores with the other datasets highlights the importance of using real MDS datasets for this task.

To get better insights into the individual contributions of the sources, we analyze their variance, where a value of 0 indicates that all sources contribute equally to the summary. Table~\ref{tab:unique_variance} demonstrates that variance, and hence, the variability of their individual contributions, increases with the number of sources. Furthermore, we compute the frequency at which each source contributes the most with their unique information (Figure~\ref{fig:top_unique_plot}). Interestingly, we observe that the first three sources consistently contribute the most in all datasets, regardless of the number of sources.  Similarly, \citet{wolhandler-etal-2022-multi} report that the information in multi-document summaries is often covered by a single source document. Our results also highlight a strong bias towards the order of the sources when summarizing multiple documents, consistently observed across the datasets.
%This indicates that there is a bias towards the order of the sources when summarizing multiple documents. 

\begin{table}%[hbt]
\small
\centering
\resizebox{\columnwidth}{!}{
\begin{tabular}{lrrrr}
%\hline 
\toprule
%\midrule
\textbf{Answer} & \textbf{Union} & \textbf{Synergy} & \textbf{Red.} & \textbf{Unique}\\
\midrule
Unrelated & 0.04 ($\pm$0.3) & 0.24 ($\pm$0.5) & 0.01 ($\pm$0.2) & 0.03 ($\pm$0.1)\\
Incorrect & 0.05 ($\pm$0.3) & 0.26 ($\pm$0.5) & 0.02 ($\pm$0.2) & 0.03 ($\pm$0.1)\\
Correct & 0.05 ($\pm$0.3) & 0.28 ($\pm$0.5) & 0.02 ($\pm$0.2) & 0.04 ($\pm$0.1)\\

%Unrelated & 0.0 / 0.30  & 0.07 / 0.50 & 0.0 / 0.16 & 0.0 / 0.11\\
%Incorrect & 0.0 / 0.35 & 0.10 / 0.50 & 0.0 / 0.21 & 0.0 / 0.12\\
%Correct & 0.0 / 0.33 & 0.11 / 0.49 & 0.0 / 0.18 & 0.0 / 0.12\\
%\midrule
\bottomrule
\end{tabular}
}
\caption{\label{tab:synergy} \spider~scores on MultiRC as MDS data (mean and standard deviation).}% See all statistics in Table~\ref{tab:apx_synergy}.}
\end{table}

\begin{table*}
\small
\centering
\begin{tabular}{p{15.4cm}l}
\textbf{Sources}\\
\toprule 
The story revolves around an upright and principled Police Officer, A.C.P. Ramakant Chaudhary whose eldest son Vikas is killed in a pre-planned accident.\\
\cmidrule(lr){1-1}
The day comes when Vishal confronts Baba Khan and Manna Shetty which leads to tension and gory situation for the A.C.P., as the ganglords threaten to eliminate the A.C.P. as well as his wife Revati and son Vishal.\\
%\hline
%\bottomrule
\end{tabular}
%\end{table*}
%\begin{table*}[h]
%\small
%\centering
\begin{tabular}{llp{11.1cm}l}
%\hline 
\midrule
\textbf{Summary} & Correct Answer & What is the name of Revati’s husband? Ramakant Chaudhary \\
%\midrule
%\hline
\textbf{Summary} & Incorrect Answer& What is the name of Revati’s husband? Baba Khan \\
%\midrule
%\hline
\textbf{Summary} & Unrelated Q\&A & How many people come to comfort the baby? 2 \\
\bottomrule
\end{tabular}
\caption{\label{tab:multirc_example} Examples of sources-summary instances of MultiRC as a MDS dataset. Each sentence represents an independent source and the summary is the concatenation of the question and answer. For each question in the dataset, we generate an unrelated instance consisting of a question and answer from a different paragraph.}
\end{table*}
\section{Measuring Synergistic Information}
Since synergy is negligible in the analyzed MDS summaries (see Section~\ref{sec:analysis}), we perform an additional experiment to assess whether our approach can measure synergistic information using the MultiRC dataset~\cite{khashabi-etal-2018-looking}. Specifically, MultiRC is a reading comprehension dataset consisting of multi-sentence paragraphs and multiple-choice questions, which also specifies the sentences that are required to answer each question. Most importantly, the dataset ensures that correct answers can only be derived by considering multiple sentences jointly. That is, synergistic information should be prominent in correct answers.

%Additionally, the dataset was constructed such that the answers are not necessarily part of the source text, so lexical matching approaches are not sufficient to identify correct answers.

To apply our PID approach, we transform MultiRC into a MDS dataset, where each instance comprises the set of sentences (sources) required to answer a question, and a question-answer pair concatenated into a single sentence (summary). Although both correct and incorrect answers share the same question, the former should result in higher synergy than the latter. We also generate unrelated instances for each set of source sentences, where the question-answer pairs are randomly sampled from a different paragraph (see Figure~\ref{tab:multirc_example}).

Table~\ref{tab:synergy} confirms that synergistic information is the dominant information component and correct answers achieve the highest synergy. However, synergy is also present in unrelated instances. Given that synergy represents new information, this raises the question whether it could also be an indicator of hallucination.~This is an interesting research direction for future work, since more experiments are needed to support this hypothesis.
%In contrast to the MDS analysis in Section~\ref{sec:analysis}, synergy carries most of the information and the values of redundancy and unique information are almost negligible. 

%split: -1
%Union -- Mean: 0.03921662350123937, Median: 0.0, Std_dev: 0.29483572857849916
%Synergy -- Mean: 0.24065503908252206, Median: 0.06820231818440803, Std_dev: 0.4998334352117261
%Redundancy -- Mean: 0.01088461266496499, Median: 0.0, Std_dev: 0.15870265125555855
%Unique total -- Mean: 0.013135704939153139, Median: 0.0, Std_dev: 0.1097113035401705

%split: 0
%Union -- Mean: 0.048856648077870955, Median: 0.0, Std_dev: 0.3469560199846484
%Synergy -- Mean: 0.26376489788730006, Median: 0.09637401631301315, Std_dev: 0.4973348918333331
%Redundancy -- Mean: 0.017044805627521774, Median: 0.0, Std_dev: 0.20935287354861337
%Unique total -- Mean: 0.015064722660355197, Median: 0.0, Std_dev: 0.11962659023331192

%split: 1
%Union -- Mean: 0.05355881767009425, Median: 0.0, Std_dev: 0.33178736008016113
%Synergy -- Mean: 0.27506728740291103, Median: 0.10805408420132735, Std_dev: 0.47846692110801053
%Redundancy -- Mean: 0.0186650128953165, Median: 0.0, Std_dev: 0.17552092373039707
%Unique total -- Mean: 0.016481776587275662, Median: 0.0, Std_dev: 0.12100471858671358

\section{Conclusion}
%% Maybe the conclusion is a bit long, just visually
We characterize the information present in human-written multi-document summaries to get insights into what information comprises a high-quality summary.~In particular, we suggest to decompose the mutual information that the source documents provide about a summary% into information present in all sources (redundancy), at least one source (union), only one source (unique), and new information that results form considering all sources jointly (synergy) 
~and propose \spider, a novel approach to quantify such information % in multi-document summaries
using partial information decomposition.~We then analyze human-written summaries from widely used MDS datasets.~The results reveal that redundancy decreases, whereas unique information increases with the number of sources. Furthermore, the order of the documents has an impact on the summarization process, as the first three documents contribute the most in terms of unique information.

\section*{Limitations}
Some limitations of our work can be traced to the beam-search-based approximation caused by the intractable sentence search space, or due to using sentences as our base unit of information. We also note that despite the remarkable recent advances, large language models' probability distributions possibly diverge from the true underlying distribution. Their approximation is, however, continuously improving and future models can directly be substituted into our method. 

% beam search
% sentences as unit of information

% the probability distribution of a LM is maybe not necessarily a perfect approximation of a probability distribution for information theory (this may be a limitation to this framework

\section*{Ethics Statement}
From an ethical perspective, it is important to underline the importance of transparency when using language models, as they are becoming nearly indistinguishable from human writers. We advocate for clear transparency when they are used. Our work promotes interpretability in the space of multi-document summarization, and we hope both interpretability and transparency will be cornerstones for future work in the field. 

In all our experiments, we rigorously follow the ACL Code of Ethics, using pre-existing open-source benchmark datasets where privacy concerns were already addressed by the respective authors.

% Throughout our experiments, we strictly adhere to the ACL Code of Ethics. Since we used already established open-source benchmark datasets, the concern of privacy does not apply. Furthermore, since no additional data was collected or stored. 

\section*{Acknowledgements}
This project is supported by Ringier, TX Group, NZZ, SRG, VSM, viscom, and the ETH Zurich Foundation.

\bibliography{acl_latex}

\end{document}